\documentclass[letterpaper, 10 pt, conference]{ieeeconf}  

\IEEEoverridecommandlockouts                              

\usepackage[T1]{fontenc}
\usepackage{aecompl}

\usepackage{times}
\usepackage{epsfig}
\usepackage{graphicx}
\usepackage{amsmath}
\usepackage{amssymb}
\usepackage{algorithm}
\usepackage{algorithmic}
\usepackage{multirow}
\usepackage{subfigure}
\usepackage[table]{xcolor}
\usepackage{caption}

\newcommand{\ie}[1]{\emph{i.e.~}}
\newcommand{\eg}[1]{\emph{e.g.~}}
\newcommand{\etal}{\emph{et~al~}}
\usepackage[breaklinks=true,letterpaper=true,colorlinks,bookmarks=false]{hyperref}

\title{\LARGE \bf
Adversarial Attacks on Monocular Depth Estimation
}

\author{
Ziqi Zhang$^{\ddag}$, 
Xinge Zhu$^{\dag}$, 
Yingwei Li$^{\S}$, 
Xiangqun Chen$^{\ddag}$, 
Yao Guo$^{\ddag}$ 
\\ $^{\ddag}$Peking University,~~~ $^{\dag}$The Chinese University of Hong Kong,~~~ $^{\S}$Johns Hopkins University
}

\begin{document}

\maketitle
\thispagestyle{empty}
\pagestyle{empty}

\begin{abstract}

Recent advances of deep learning have brought exceptional performance on many computer vision tasks such as semantic segmentation and depth estimation.
However, the vulnerability of deep neural networks towards adversarial examples have caused grave concerns for real-world deployment.
In this paper, we present a systematic study of adversarial attacks on monocular depth estimation, an important task of 3D scene understanding in scenarios such as autonomous driving and robot navigation.
In order to understand the impact of adversarial attacks on depth estimation, we first define a taxonomy of different attack scenarios for depth estimation, including \textit{non-targeted attacks}, \textit{targeted attacks} and \textit{universal attacks}.
We then adapt several state-of-the-art attack methods for classification on the field of depth estimation.
Besides, multi-task attacks are introduced to further improve the attack performance for universal attacks.
Experimental results show that it is possible to generate significant errors on depth estimation.
In particular, we demonstrate that our methods can conduct targeted attacks on given objects (such as a car), resulting in depth estimation $3$-$4\times$ away from the ground truth (e.g., from 20m to 80m).

\end{abstract}

\section{Introduction}

Monocular depth prediction, \ie~predicting the per-pixel distances to the camera, is a key task for 3D scene understanding.
Learning 3D scene geometries has many applications such as robot assisted surgery, robot navigation and autonomous driving ~\cite{coupete2015gesture,desouza2002vision,stoyanov2010real,Uhrig2017THREEDV,ma2019trafficpredict}.
With its wide-spread applications, more and more works have significantly promoted the monocular depth estimation performance using DCNN-based models~\cite{eigen2015predicting,kim2016unified,kuznietsov2017semi,laina2016deeper,liu2016learning,roy2016monocular,wang2015towards}.

Meanwhile, the advances in deep learning have largely pushed forward the state-of-the-art in various tasks in computer vision, such as image classification and semantic segmentation.
However, several studies~\cite{goodfellow2014explaining,szegedy2013intriguing} have demonstrated the vulnerability of deep learning based methods to deliberately generated adversarial examples, bringing reliability concerns to the applications in many safety-critical domains, such as autonomous driving and video surveillance.
Hence, more and more work~\cite{carlini2016towards,kurakin2016adversarial} about adversarial attacks have been proposed to evaluate the robustness and reliability of neural network models before they are deployed in the real world.

\begin{figure}[t]

	\centering
    \subfigure[{\scriptsize Original mean object depth is 8.2m}]{

    \label{fig:rgb}
    \includegraphics[width=0.475\linewidth]{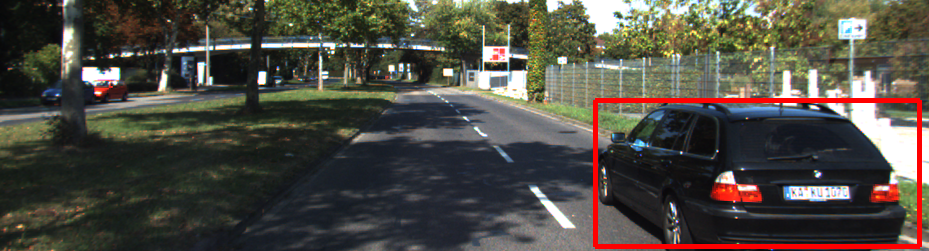}
    \includegraphics[width=0.475\linewidth]{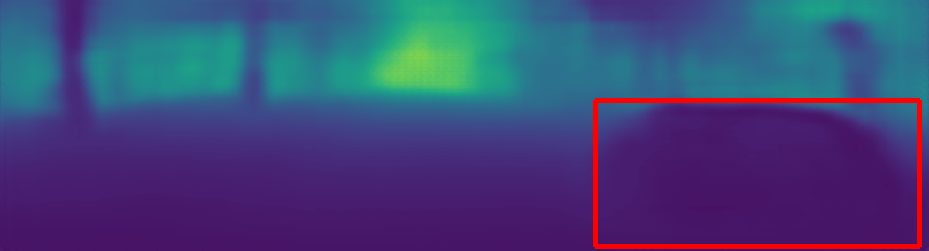}
    }

    \vspace{-0.5\baselineskip}
    \subfigure[{\scriptsize Adversarial mean object depth is 3.7m ($0.5\times$)}]{
    \label{fig:rgb}
    \includegraphics[width=0.48\linewidth]{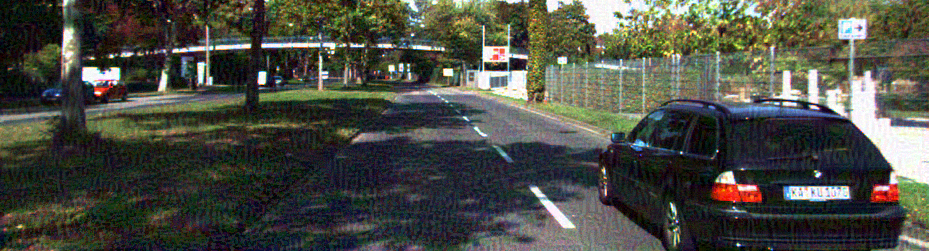}
    \label{fig:double-pong}
    \includegraphics[width=0.48\linewidth]{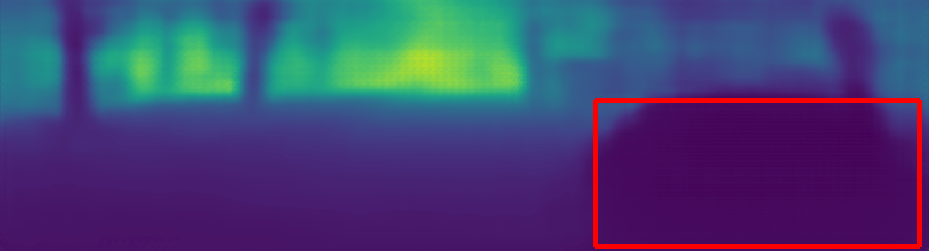}}

    \vspace{-0.5\baselineskip}
    \subfigure[{\scriptsize Adversarial mean object depth is 35.5m ($4.3\times$)}]{
    \label{fig:double-pong}
    \includegraphics[width=0.48\linewidth]{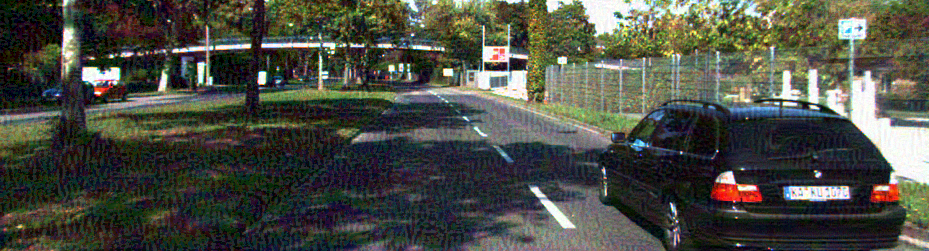}
    \includegraphics[width=0.48\linewidth]{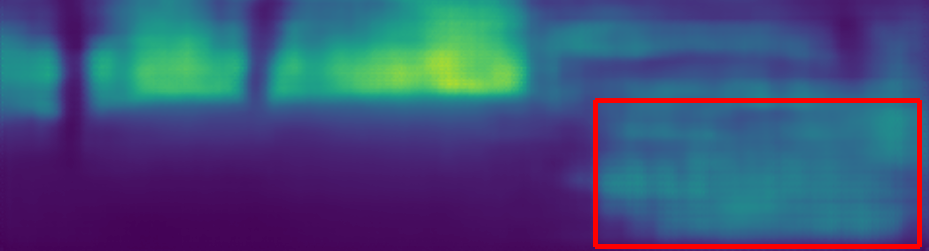}
    }

    \vspace{-0.5\baselineskip}
    \subfigure[{\scriptsize Adversarial mean object depth is 68.6m ($8.4\times$)}]{
    \label{fig:double-pong}
    \includegraphics[width=0.48\linewidth]{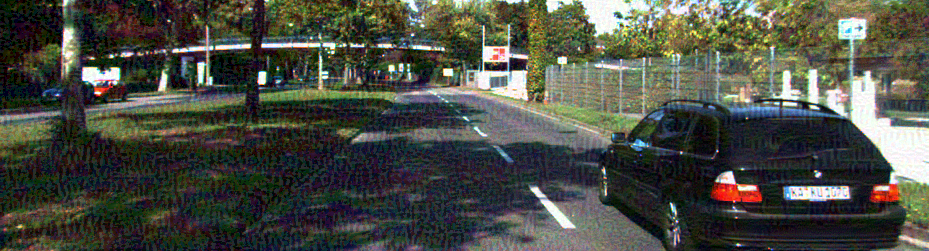}
    \includegraphics[width=0.48\linewidth]{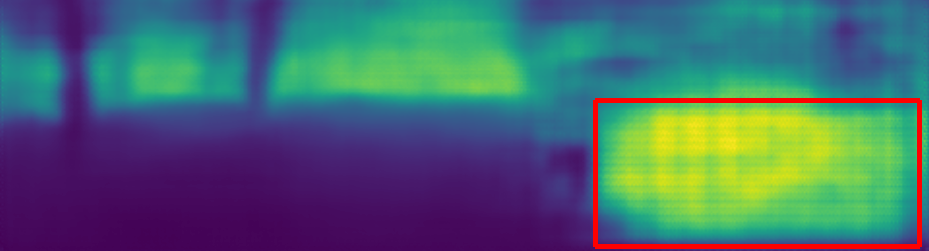}
    }


    \caption{An illustration of the \textbf{targeted attack} on monocular depth estimation. The goal of this attack is to mislead the depth of specific objects (the black car in the red box in this case).
    The left column contains RGB images and the right column includes the corresponding depth prediction results.
    The first row shows the original images while three different adversarial examples are shown in the following rows.
    In the depth prediction image, bright yellow represents large depth value while dark blue represents small value.
    The depth estimation results are distorted up to $8\times$ from the original prediction.
    Best viewed in color.
    }
    \vspace{-3ex}
    \label{fig:intro:traser}
\end{figure}

Recent studies on adversarial attacks have served as the foundation of adversarial training, \ie~a common technique to enhance model robustness~\cite{athalye2018obfuscated, madry2017towards,   shaham2018understanding, sinha2017certifying}.
Specifically, adversarial examples generated from the attack methods are combined into training data.
Models trained with the mixture of clean data and adversarial data thus become more robust to adversarial attacks.
Similar to attacks on classification, a systematic study of attack on depth estimation could also strengthen the robustness of depth models.

Although a number of attack methods~\cite{szegedy2013intriguing, kurakin2016scale, dong2017boosting, carlini2017adversarial,metzen2017universal, goodfellow2014explaining,kurakin2016adversarial} have been proposed, they mainly focus on classification~\cite{dong2017boosting, szegedy2013intriguing, kurakin2016scale} or segmentation~\cite{metzen2017universal, arnab2017robustness,sun2019not}.
Both tasks aim to pick a label among a fixed number of categories, so these techniques are not designed specifically to attack tasks such as monocular depth estimation, which aims to regress a precise value for each pixel.

In order to investigate the problem of adversarial attacks on monocular depth estimation, we first define a taxonomy of different attack scenarios for depth estimation, including \textit{non-targeted attacks }(for a specific image), \textit{targeted attacks} (for a specific object in an image) and \textit{universal attacks} (for an arbitrary image).
We then adapted several state-of-the-art attack methods to perform these attacks.

Moreover, we introduce a new multi-task attack strategy to improve the performance in the universal attack scenario.
Multi-task strategies have been widely applied~\cite{zhu2018penalizing,chen2017gradnorm,kokkinos2017ubernet,zhu2019adapting} in supervised learning, where various supervision signals provide a more thorough understanding.
In our setting, the segmentation task and depth estimation form the multi-task attack strategy (\ie, attacking both tasks simultaneously), where the high-level task (semantic segmentation) and low-level task (depth estimation) could offer complementary information to further boost the attacking performance for universal attacks.

We have performed extensive experiments on the popular KITTI dataset~\cite{geiger2013vision}.
Results demonstrate that the average depth estimation errors (\ie~RMSE) can be distorted up to $10\times$ compared with the original prediction, which means that the adversarial attacks are real and practical. For targeted attacks, the distance estimation results can be distorted up to $3$-$4\times$ (e.g., from 20m to 80m) for given objects. Fig.~\ref{fig:intro:traser} displays an example. In addition, our proposed multi-task method achieves a better performance in the universal attacks, compared to single-task setting.

The main contributions of this work are as follows.
\begin{itemize}
	\item This paper presents a systematic study of adversarial attacks on the task of monocular depth estimation.
    We define and benchmark various experimental settings for depth attacks, including non-targeted attacks, targeted attacks and universal attacks.
    \item We have adapted several state-of-the-art adversarial attack methods for classification to perform attacks on depth estimation.
    A new attack strategy, multi-task attack, is introduced to enforce attacking performance on single task with the support of auxiliary task by enriching supervision signals.
    \item Experimental results demonstrate that adversarial attacks on depth estimation is a real threat as we are able to generate significant errors in the estimation results.
    In particular, we are able to attack specific objects such as pedestrians and cars, bringing out much larger results than the clean estimation.
    In addition, our proposed multi-task attack strategy achieves better performance than existing methods.
    \end{itemize}


%
%

\section{Related Work}

\subsection{Monocular Depth Estimation}
Monocular depth estimation plays an important role for understanding spatial structures from 2D images.
Traditional methods use triangulation to compute spatial position of each point corresponding in stereo images.
Saxena~\etal~\cite{saxena2009make3d} proposed a supervised learning approach leveraging local- and global-features. After that, various of methods based on hand-crafted features have been proposed \cite{hane2015direction, abadi2016tensorflow, ranftl2016dense,xu2019depth}.

\vspace{1ex}\textbf{DCNN based techniques}. Eigen~\etal first predicted dense depth estimation with deep neural networks~\cite{eigen2014depth}.
Supervised deep learning-based approaches thus advanced the state of the art with different techniques.
Xie~\etal utilized the skip-connection strategy to fuse low-resolution feature maps and high-resolution feature maps~\cite{xie2016deep3d}.
Laina~\etal proposed a fully convolutional architecture with residual learning and an up-sampling module~\cite{laina2016deeper}.
Fu~\etal proposed an ordinal regression loss
to recast depth network learning as a classification problem~\cite{fu2018deep}.
In some work, CRF was integrated into deep architectures as well~\cite{liu2015deep, wang2015towards, xu2018structured} .

\vspace{1ex}\textbf{Multi-task strategy}. Recent work further demonstrated that depth estimation can be learned in a multi-task setting.
Xu~\etal proposed to simultaneously solve the problem of depth estimation and scene parsing in a joint CNN~\cite{xu2018PAD-Net}.
Zhang~\etal proposed a task-recursive-learning framework for semantic segmentation and depth estimation~\cite{zhang2018joint}.


\subsection{Adversarial Attacks}

\vspace{1ex}\textbf{White-box and black-box attacks}.
White-box attack means that attackers have the access of the target model.
Szegedy~\etal suggested that adding human imperceptible perturbations leads deep neural networks to wrong predictions~\cite{szegedy2013intriguing}.
Later, utilizing the linear property of neural network, a fast attack method was developed and named as Fast Gradient Sign Method~\cite{goodfellow2014explaining}.
Later, an iteration based method~\cite{kurakin2016adversarial} was built to generate stronger adversarial examples.
On the contrary, in black-box attack, attackers cannot access the target model.
People developed a variety of methods to attack a black-box model, including query-based~\cite{bhagoji2018practical,chen2017zoo}, decision-based~\cite{guo2018low}, and transfer-based methods~\cite{baluja2017adversarial,dong2017boosting,poursaeed2017generative,xiao2018generating,xie2018improving,Zhou_2018_ECCV}.
Several work studied the adversarial attacks on semantic segmentation~\cite{arnab2017robustness, poursaeed2018generative, metzen2017universal}.





\vspace{1ex}\textbf{Per-image and universal attacks}.
Per-image attacks mean to  generate an adversarial example for each given image specifically.
Oppositely, universal attacks generates a universal perturbation that can be directly added to any test image to fool the classifier.
Moosavi-Dezfooli~\etal suggested the existence of universal adversarial perturbations and generated them by iteratively optimizing the per-instance adversarial loss~\cite{moosavi2016universal}.
Shafahi~\etal developed a much faster universal adversarial attack method~\cite{shafahi2018universal}.
Mopuri~\etal generated data-independent universal perturbation by maximizing spurious activations at each layer~\cite{mopuri-bmvc-2017}.
Metzen generated universal adversarial perturbations against semantic segmentation~\cite{metzen2017universal}.

To the best of our knowledge, no existing work have studied adversarial attacks on tasks such as monocular depth estimation.

\subsection{Monocular Depth Estimation and Adversarial Attacks}
Some work relates to both monocular depth estimation and adversarial attacks.
Van~\etal~\cite{van2019neural} crafted some special cases to study the internal mechanism of how networks see depth from an image. The difference between this work and \cite{van2019neural} is that this paper studies human-undetectable adversarial perturbation while Van~\etal constructs obvious fake images that can be recognized from one sight.
Mopuri~\etal~\cite{mopuri-bmvc-2017} proposed a data-free method to craft universal adversarial examples for a given CNN. Although their method is effective on depth estimation and semantic segmentation respectively, it aims at a single network and can not generate universal examples that can attack both tasks from a different structure at the same time. On the other hand, the technique described in Section~\ref{sec:method:universal} generates universal examples that attack two tasks from two different networks.
Hu~\etal~\cite{hu2019analysis} studied the white-box adversarial attacks (I-FGSM) on depth estimation in an indoor situation and proposed to defense by saliency map. By comparison, this paper studies a more dangerous situation: the depth of a certain object can be manipulated in autonomous driving. Besides, this paper utilizes multi-task (depth estimation and semantic segmentation) to attack better while Hu~\etal employs multi-task (depth estimation and saliency map) to defense.

\section{Methodology}

In this section, we first formulate the problem of adversarial attacks on depth estimation in Section~\ref{sec:method:formulation}. Then Section~\ref{sec:method:background} presents three state-of-the-art methods we adapted from classification. Finally, multi-task strategy is illustrated in Section~\ref{sec:method:universal} .

\subsection{Problem Formulation}
\label{sec:method:formulation}
Let $\bf{x} \in \mathbb{R}^{3\times h\times w}$ be an input image and $\bf{y}^{\text{true}} \in \mathbb{R}^{h\times w}$ be the ground truth depth image.
$f$ represents deep neural network parameterized by $\theta$ and $L(f(\bf{x};\theta), \bf{y}^{\text{true}})$ is the loss function.
We denote $\xi$ as the adversarial perturbation and let $\bf{x}^\text{adv}=\bf{x}+\xi$ be the corresponding adversarial example.
To make the adversarial example imperceptible, we restrict the perturbation $\|\xi\|_{\infty}<\epsilon$, where $\epsilon$ is a given perturbation constraint. More specifically, this paper considers three different types of depth attacks: \emph{non-targeted attacks}, \emph{targeted attacks} and \emph{universal attacks}.

\vspace{1ex}
 \textbf{Non-targeted attacks.}
Its goal is to maximize the prediction error for a given image, such that the original model would predict incorrectly (output incorrect depth values), which is the most typical attack type.
It maximizes a loss function as described by Eq.~\ref{equ:non-target}:
\begin{gather}
	\max_{{\bf{x}}^{\text{adv}}-{\bf{x}}} L(f({\bf{x}}^{\text{adv}};\theta),{\bf{y}}^{\text{true}}) \;\; s.t. \; \|{\bf{x}}-{\bf{x}}^{\text{adv}}\|_{\infty}<\epsilon. \label{equ:non-target}
\end{gather}

\vspace{1ex}
 \textbf{Targeted attacks.}
Taking into consideration the characteristics of depth estimation, this attack aims to mislead the model to produce incorrect depth estimation for specific (masked) objects towards a predefined depth value, whose objective is formulated in Eq.~\ref{equ:mask-target}:
\begin{gather}
	\min_{{\bf{x}}^{\text{adv}}-{\bf{x}}} L(f({\bf{x}}^{\text{adv}} ;\theta), C \cdot M + {\bf{y}}^\text{true} \odot (1-M)) \;\; \nonumber \\
	s.t. \; \|{\bf{x}}-{\bf{x}}^{\text{adv}}\|_{\infty}<\epsilon, \label{equ:mask-target}
\end{gather}
where $M$ is a binary object mask and $C$ is a predefined depth value. The goal is to cause mis-estimation of certain objects while preserving the other parts.
For example, a targeted attack could induce an autonomous driving car to predict a rider ahead to be farther away.

\vspace{1ex}
 \textbf{Universal attacks.}
This third attack type aims to train a universal adversarial perturbation (UAP) that can be added to a broad class of images.
The training goal is described in Eq.~\ref{equ:universal}, where $N$ is the training image number.
It empowers attackers who cannot generate per-instance adversarial examples on the go with an image-agnostic perturbation.
\begin{gather}
	\max_{{\bf{x}}^{\text{adv}}-{\bf{x}}} \frac{1}{N} \sum_{i=0}^{N}L(f({\bf{x}}^{\text{adv}}_{\text{i}};\theta),{\bf{y}}^{\text{true}}_{\text{i}}) \nonumber \\
	s.t.\; \|\bf{x}_{\text{i}}-\bf{x}^{\text{adv}}_{\text{i}}\|_{\infty}<\epsilon \label{equ:universal}
\end{gather}

\subsection{Adapting Existing Attacking Methods} \label{sec:method:background}
In order to reveal the impact of adversarial attacks on depth estimation, we first adapted three state-of-the-art attack methods to attack depth estimation.
These tasks include Fast Gradient Sign Method (FGSM)~\cite{goodfellow2014explaining}, Iterative FGSM (I-FGSM)~\cite{kurakin2016adversarial}, and Momentum I-FGSM (MI-FGSM)~\cite{dong2017boosting}. With the loss functions defined earlier, we implement each of the methods as follow.

\vspace{1ex}
 \textbf{FGSM.}~~It computes the direction of the loss gradient, and then adds the max possible perturbations on the original image, by
\begin{equation}
\vspace{-0.7ex}
{\bf{x}}^{\text{adv}} = {\bf{x}} + \epsilon\cdot\text{sign}\left(\nabla_{{\bf{x}}} L(f({\bf{x}};\theta), {\bf{y}}^{\text{true}})\right)
\end{equation}
where sign$(\cdot)$ denotes the sign function.

\vspace{1ex}
 \textbf{I-FGSM.}~~It initializes an adversarial example ${\bf{x}}_{0}^{\text{adv}} = {\bf{x}}$ and then iteratively updates it by
\begin{equation}
{\bf{x}}_\text{t+1}^\text{adv} = \text{Clip}_{\bf{x}}^{\epsilon} \{{\bf{x}}_\text{t}^{\text{adv}} + \alpha \cdot \text{sign}\left(\nabla_{\bf{x}} L(f({\bf{x}}_\text{t}^{\text{adv}};\theta), {\bf{y}}^{\text{true}})\right)\},
\end{equation}
The clip function $\text{Clip}_{\bf{x}}^{\epsilon}$ ensures the generated adversarial example is within the $\epsilon$-ball of the original image $x$ with ground-truth ${\bf{y}}^{\text{true}}$.

\vspace{1ex}
 \textbf{MI-FSGM.}~~This method integrates the momentum term into the attack to stabilize update directions and escape from poor local maxima. At the $t^\text{th}$ iteration, the accumulated gradient is 
\begin{equation}
{\bf{g}}_\text{t+1} = \mu \cdot {\bf{g}}_\text{t} + \frac{\nabla_{\bf{x}} L(f({\bf{x}};\theta), {\bf{y}}^{\text{true}})}{||\nabla_{\bf{x}} L(f({\bf{x}};\theta), {\bf{y}}^{\text{true}})||_1},
\end{equation}
where $\mu$ is the momentum decay factor. The sign of $g_{n+1}$ is then used to generate the adversarial example, by
\begin{equation}
{\bf{x}}_\text{t+1}^{\text{adv}} = \text{Clip}_{\bf{x}}^{\epsilon}\{{\bf{x}}_\text{t}^{\text{adv}} + \alpha \cdot \text{sign}({\bf{g}}_\text{t+1})\}.
\label{equ:mi_fgsm_update}
\end{equation}

\subsection{Multi-Task Strategy for Universal Attacks}
\label{sec:method:universal}

Typically, universal attacks are  not as effective as non-targeted or targeted attacks in term of the increased error, as they often attack a set of images simultaneously.
To further improve the performance of universal attacks, we introduce the multi-task attack strategy.
Specifically,
In our setting, both segmentation and depth estimation tasks are available. The proposed strategy aims to generate the universal adversarial perturbation to attack both tasks simultaneously.
The conjugation of low-level information provided by depth estimation and semantic information from segmentation could offer more complementary signals than the single task, thus boosting the performance of depth attacking.
%
%
%

Due to the lack of ground truth (no depth ground truth and large-scale segmentation labels coexist simultaneously in one dataset), we conduct the non-targeted attack to depth estimation and use the least-likely method (LLM)~\cite{kurakin2016adversarial} to perform the segmentation task.  LLM takes the least likely label $\mathbf{y}^\text{LL}=\arg \min_\mathbf{y} \text{p}(\mathbf{y}|\mathbf{x})$ as the attack target and increases the prediction probability of that label as described in Eq.~\ref{equ:LLM}.
\begin{gather}
	\min_{{\bf{x}}^{\text{adv}}-{\bf{x}}} L(f({\bf{x}}^{\text{adv}};\theta),{\bf{y}}^{\text{LL}}) \;\; s.t. \; \|{\bf{x}}-{\bf{x}}^{\text{adv}}\|_{\infty}<\epsilon \label{equ:LLM}
\end{gather}

We utilize MI-FGSM to illustrate the generation of universal adversarial perturbation with the multi-task strategy. Note MI-FGSM can be replaced by any method in section~\ref{sec:method:background} easily.
To compute the adversarial perturbation for a minibatch, we use $\overline{L}=\mathbb{E}_{\bf{x}\in \text{B}}L(f(\bf{x}, \theta), \mathbf{y})$ as the target loss function, where $\text{B}$ represents a minibatch.
In specific, we define $\overline{L}_\text{t}^\text{depth}=\mathbb{E}_{\bf{x}\in \text{B}_\text{t}}L^\text{depth}(f(\bf{x};\theta),\mathbf{y}^\text{true})$ and $\overline{L}_\text{t}^\text{semantic}=\mathbb{E}_{\bf{x}\in \text{B}_\text{t}}L^\text{semantic}(f({\bf{x}},\theta), \mathbf{y}^\text{LL})$.
 To combine information from both tasks, we compute the averaged loss~$L_\text{t}$ as Eq.~\ref{equ:multitask_loss} and averaged gradient~$\overline{\bf{g}}_\text{t}$as Eq.~\ref{equ:multitask}, where $w_{\text{depth}}$ and $w_{\text{semantic}}$ are predefined weights for each task. 

 \begin{equation}
	L_\text{t}=w_{\text{depth}} \cdot |\overline{L}_\text{t}^{\text{depth}}| + w_{\text{semantic}} \cdot |\overline{L}_\text{t}^{\text{semantic}}|
	\label{equ:multitask_loss}
	\vspace{1ex}
\end{equation}

\begin{equation}
	\overline{\bf{g}}_\text{t}=w_{\text{depth}} \cdot \frac{\nabla_{\bf{x}} \overline{L}_\text{t}^\text{depth}}{\|\nabla_{\bf{x}} \overline{L}_\text{t}^\text{depth}\|_1} + w_{\text{semantic}} \cdot \frac{\nabla_{\bf{x}} \overline{L}_\text{t}^\text{semantic}}{\|\nabla_{\bf{x}} \overline{L}_\text{t}^\text{semantic}\|_1}
	\label{equ:multitask}
	\vspace{1ex}
\end{equation}
 Our multi-task universal attack algorithm is summarized in Algorithm~\ref{multitask}.

In our implementation, two tasks are involved in the multi-task attack and note that it is easy to extend this strategy to more tasks to take advantage of more supervision signals.

\begin{algorithm}[t]
\caption{Multi-Task Universal Perturbation Generation}
\label{multitask}
\begin{algorithmic}[1]
\REQUIRE Training samples $\bf{X}$,  perturbation bound $\epsilon$, learning  rate $\gamma$, momentum $\mu$, epoch number $\text{N}_\text{ep}$ and iteration number $\text{T}$
\REQUIRE Loss function of the two tasks $L^{\text{depth}}$ and $L^{\text{semantic}}$  and weights of the two tasks $w_{\text{depth}}$ and $w_{\text{semantic}}$.

\STATE Randomly initialize $\delta$
\FOR{epoch~$ =1 \ldots \text{N}_\text{ep}$}
    \FOR{miniatch $\text{B} \subset \text{X}$}
    	\STATE Initialize $\text{B}_0=\text{B}$, $\bf{g}_0=0$ 
    	\FOR{iteration $\text{t}=\text{0}\ldots \text{T}-1$}

    		\STATE Compute single task loss $\overline{L}_\text{t}^\text{depth}$ and $\overline{L}_\text{t}^\text{semantic}$
    		\STATE Compute single task gradient $\nabla_{\bf{x}} \overline{L}_\text{t}^\text{depth}$ and $\nabla_{\bf{x}} \overline{L}_\text{t}^\text{semantic}$
    		\STATE Compute average loss according to Eq.~\ref{equ:multitask_loss}
        	\STATE Compute average gradient according to Eq.~\ref{equ:multitask}
        	\STATE Update gradient $\bf{g}_\text{t+1}=\mu \cdot  \bf{g}_\text{t} + \overline{\bf{g}}_\text{t}$
        	\STATE Compute ${\bf{x}}_\text{t+1}$ by Eq.~\ref{equ:mi_fgsm_update} and update minibatch $\text{B}_\text{t+1}=\{{\bf{x}}_\text{t+1}\}$
        \ENDFOR
        \STATE Compute perturbation $\delta_{\text{B}}=\mathbb{E}_{{\bf{x}}_\text{T} \in \text{B}_\text{T}}({\bf{x}}_\text{T}-{\bf{x}}_0)$
        \STATE Update universal perturbation $\delta \leftarrow \delta + \gamma \cdot \delta_{\text{B}}$
    \ENDFOR
\ENDFOR
\RETURN Universal perturbation $\delta$

\end{algorithmic}

\end{algorithm}

\section{Experiments}

\begin{table*}[t]
\scriptsize
\begin{center}
\begin{tabular}{c|c|c|c|c|c||c|c|c|c}
\hline
\multirow{2}*{ } & Metric & \multicolumn{4}{c||}{RMSE $\uparrow$ (\textbf{non-targeted attack})} & \multicolumn{4}{c}{MMD $\uparrow$ (\textbf{targeted attack})} \\
\hline
 & Method & ResNet-18 & VGG-16 & ResNet-50 & ResNet-101 & ResNet-18 & VGG-16 & ResNet-50 & ResNet-101 \\
\hline
\multirow{4}*{ResNet-18} 
 & FGSM & \cellcolor{gray!25}11.54 ($2.7\times$) & 	10.30 ($2.4\times$)	& 11.02 ($2.7\times$)& \textbf{10.18} ($2.2\times$)& \cellcolor{gray!25}23.51 ($1.1\times$)& 	18.45 ($0.9\times$)& 19.34 ($0.9\times$) & 19.66 ($0.9\times$)  \\
 & I-FGSM & \cellcolor{gray!25}\textbf{35.05} ($8.3\times$)  &	10.30 ($2.4\times$)&	16.02 ($3.9\times$)& 7.95 ($1.7\times$)& \cellcolor{gray!25}72.33 ($3.5\times$) & 26.95 ($1.3\times$)&	34.42 ($1.7\times$)& 25.46 ($1.2\times$) \\
 & MI-FGSM & \cellcolor{gray!25}32.37 ($7.7\times$)	&\textbf{11.60} ($2.6\times$)&\textbf{16.70} ($4.0\times$)& 9.63 ($2.0\times$)& \cellcolor{gray!25}\textbf{72.35} ($3.5\times$)	&\textbf{27.38} ($1.3\times$)&\textbf{35.54} ($1.7\times$)& \textbf{25.57} ($1.2\times$)\\
\hline
\multirow{4}*{VGG-16} & FGSM & 8.69 ($2.1\times$) &	\cellcolor{gray!25}11.5	($2.6\times$) & 9.45 ($2.3\times$)& \textbf{9.30 ($2.0\times$)} & 22.90 ($1.1\times$) &	\cellcolor{gray!25}24.63 ($1.2\times$)& 22.98 ($1.1\times$)& 19.77 ($0.9\times$) \\
 & I-FGSM & 7.13 ($1.7\times$) & \cellcolor{gray!25}\textbf{37.40} ($8.5\times$)& 10.48 ($2.5\times$)& 6.34 ($1.3\times$)& 23.39 ($1.1\times$)& \cellcolor{gray!25}74.82 ($3.6\times$)& 25.73 ($1.2\times$) & \textbf{23.09 ($1.1\times$)} \\
 & MI-FGSM & \textbf{9.00} ($2.1\times$) & \cellcolor{gray!25}36.23 $8.2\times$)& \textbf{12.52} ($3.0\times$)& 8.26 ($1.8\times$)& \textbf{24.08} ($1.1\times$)& \cellcolor{gray!25}\textbf{76.47} ($3.7\times$)& \textbf{27.07} ($1.3\times$) & 22.76 ($1.1\times$)\\
\hline

\multirow{4}*{ResNet-50} 
 & FGSM & 10.27 ($2.4\times$)	 & 9.98	($2.3\times$)& \cellcolor{gray!25}11.58 ($2.8\times$)& \textbf{9.59 ($2.0\times$)} & 20.48	($1.0\times$) & 19.77	($1.0\times$)& \cellcolor{gray!25}25.27 ($1.2\times$) & 20.42 ($1.0\times$)\\
 & I-FGSM & 8.26 ($1.9\times$) & 8.22 ($1.9\times$)& \cellcolor{gray!25}\textbf{42.27} ($10.1\times$) & 6.42 ($1.4\times$)& 24.77 ($1.2\times$)& 24.58 ($1.2\times$) & \cellcolor{gray!25}78.72 ($3.8\times$) & 23.92 ($1.1\times$)\\
 & MI-FGSM & \textbf{10.51} ($2.5\times$) & \textbf{10.40} ($2.3\times$)& \cellcolor{gray!25}38.37 ($9.2\times$)& 8.54 ($1.8\times$)& \textbf{25.34} ($1.2\times$)& \textbf{25.21} ($1.2\times$)& \cellcolor{gray!25}\textbf{79.78} ($3.8\times$)& \textbf{23.96} ($1.1\times$) \\
\hline

\multirow{4}*{ResNet-101} & FGSM & \textbf{9.26} ($2.2\times$) & 8.69	 ($2.0\times$)& 10.22 ($2.4\times$) & \cellcolor{gray!25}12.54 ($2.7\times$)& 23.88 ($1.1\times$)& 20.45 ($1.0\times$)& 23.61 ($1.1\times$)& \cellcolor{gray!25}31.14 ($1.5\times$)\\
 & I-FGSM & 7.39 ($1.8\times$)& 7.13 ($1.6\times$)& 9.49 ($2.3\times$)& \cellcolor{gray!25}\textbf{25.25 ($5.4\times$)} & 23.87 ($1.1\times$)& 23.59 ($1.1\times$)& 26.16 ($1.3\times$)& \cellcolor{gray!25}56.35 ($2.6\times$)\\
 & MI-FGSM & 9.08 ($2.2\times$) & \textbf{9.00} ($2.0\times$)& \textbf{11.66} ($2.8\times$) & \cellcolor{gray!25}24.52 ($5.2\times$) & \textbf{24.34} ($1.2\times$)& \textbf{23.65 ($1.1\times$)} & \textbf{26.89} ($1.3\times$)& \cellcolor{gray!25}\textbf{56.65} ($2.7\times$)\\
\hline

\end{tabular}
\end{center}
\caption{The overall evaluation (RMSE represents the average depth estimation error for the whole image, while MMD indicates the mean estimated depth value of the masked target) results for adversarial attacks in white-box (shaded cells) and black-box (others) settings.
The number in the parentheses are relative to the baseline results in Table~\ref{baseline}.
A higher value indicates better attack effect.
}
\label{table:white_black_attack}
\end{table*}

\subsection{Experimental Setup}

\textbf{Dataset.} We use the KITTI depth estimation dataset~\cite{geiger2013vision}, which consists of $85,898$ training images and $6,852$ validation images, created by aggregating LiDAR scans from consecutive frames before projecting into one image.
It contains semi-dense depth ground truth of about $30\%$ annotated pixels.
We center crop every image to the size of $321 \times 929$. In the per-image attack, all images in the validation set are used.
In the universal attack, we randomly select $17,913$ images as the training set and test on the validation set.
We train universal adversarial perturbation for $2$ epochs.

\begin{figure}[h]
	\centering

    \subfigure[\scriptsize Original Image , Mean Object Depth=11.4m]{
    \label{fig:exp:specific_depth:raw}
    \includegraphics[width=0.47\linewidth]{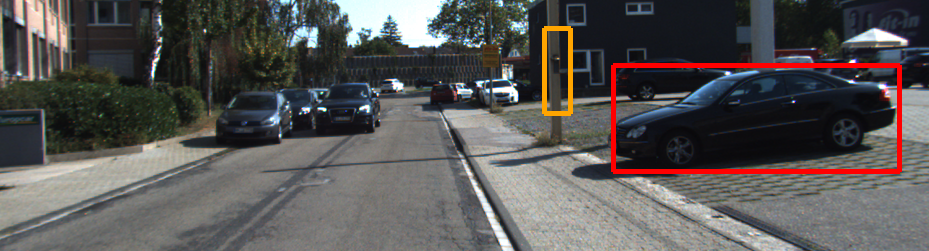}
    \includegraphics[width=0.47\linewidth]{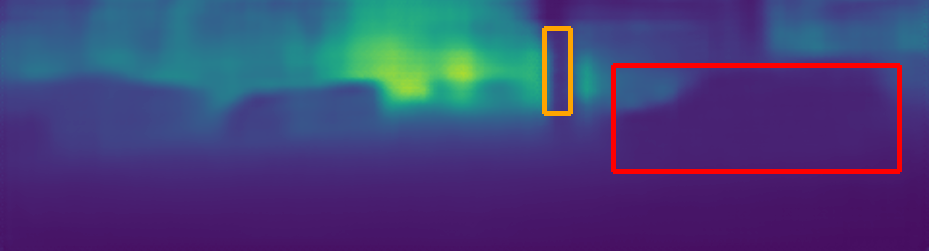}
    }

    \subfigure[\scriptsize Adversarial Example, Mean Object Depth=4.4m]{
    \label{fig:exp:specific_depth:4m}
    \includegraphics[width=0.47\linewidth]{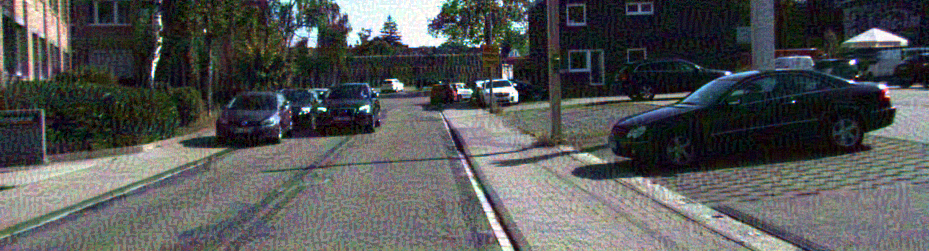}
     \includegraphics[width=0.47\linewidth]{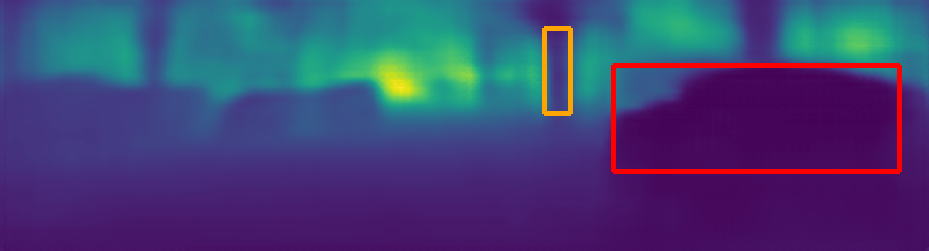}
    }

    \subfigure[\scriptsize Adversarial Example, Mean Object Depth=20.2m]{
    \label{fig:exp:specific_depth:20m}
    \includegraphics[width=0.47\linewidth]{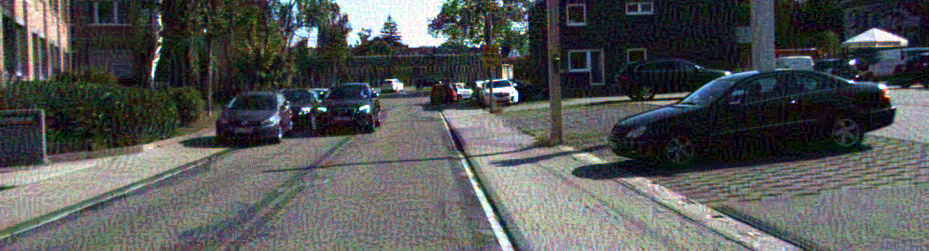}
    \includegraphics[width=0.47\linewidth]{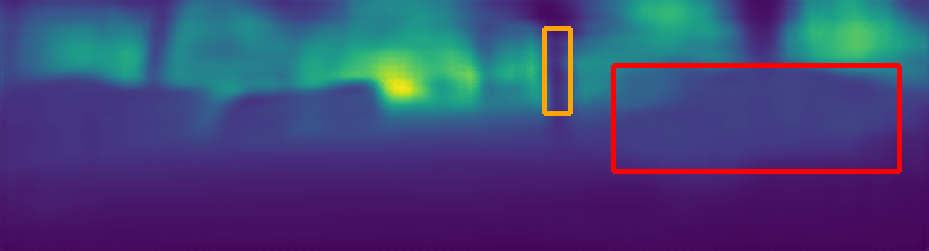}
    }

    \subfigure[\scriptsize Adversarial Example, Mean Object Depth=31.0m]{
    \label{fig:exp:specific_depth:30m}
    \includegraphics[width=0.47\linewidth]{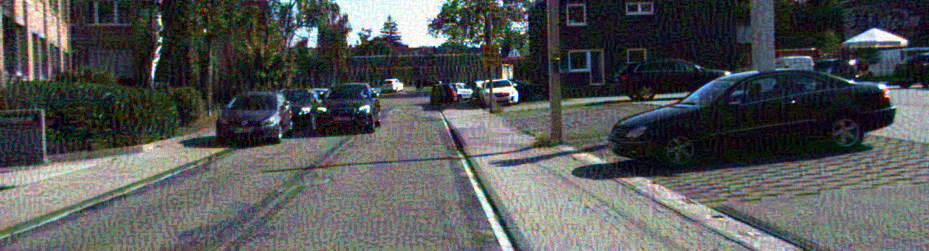}
     \includegraphics[width=0.47\linewidth]{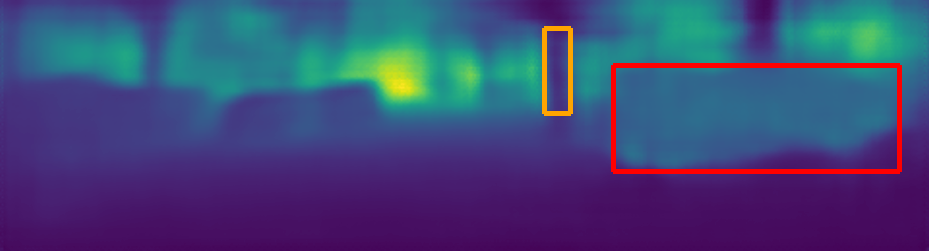}
    }


    \subfigure[\scriptsize Adversarial Example, Mean Object Depth=44.7m]{
    \includegraphics[width=0.47\linewidth]{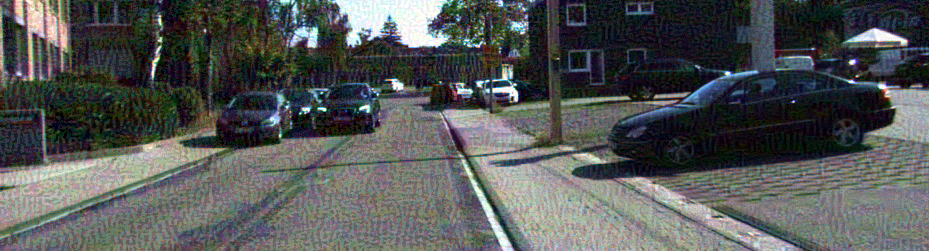}
     \includegraphics[width=0.47\linewidth]{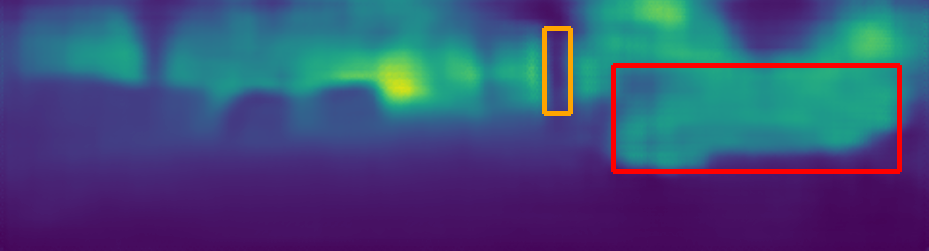}
    }

    \subfigure[\scriptsize Adversarial Example, Mean Object Depth=54.8m]{
    \includegraphics[width=0.47\linewidth]{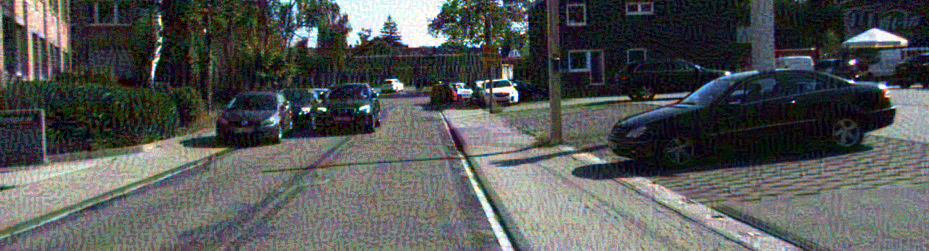}
     \includegraphics[width=0.47\linewidth]{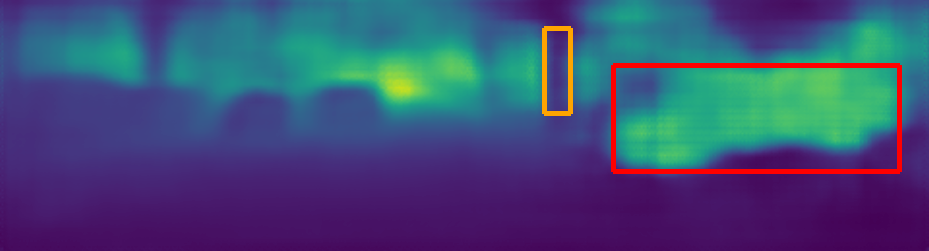}
    }

    \subfigure[\scriptsize Adversarial Example, Mean Object Depth=68.5m]{
    \label{fig:exp:specific_depth:68m}
    \includegraphics[width=0.47\linewidth]{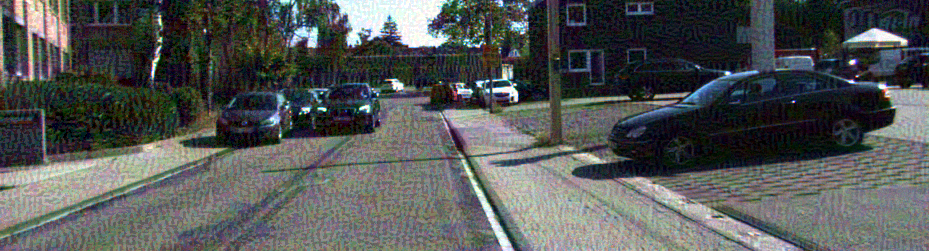}
     \includegraphics[width=0.47\linewidth]{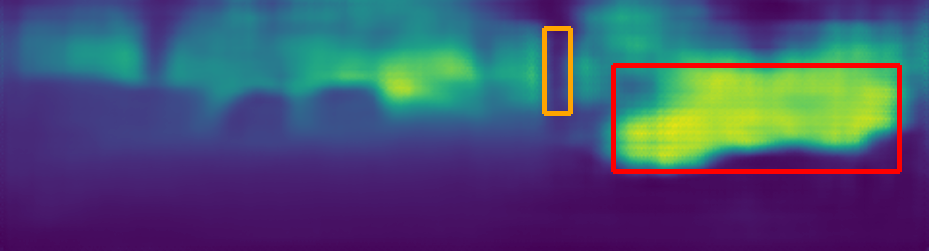}
    }

%
    \subfigure[\scriptsize Adversarial Example, Mean Object Depth=80.8m]{
    \label{fig:exp:specific_depth:80m}
    \includegraphics[width=0.47\linewidth]{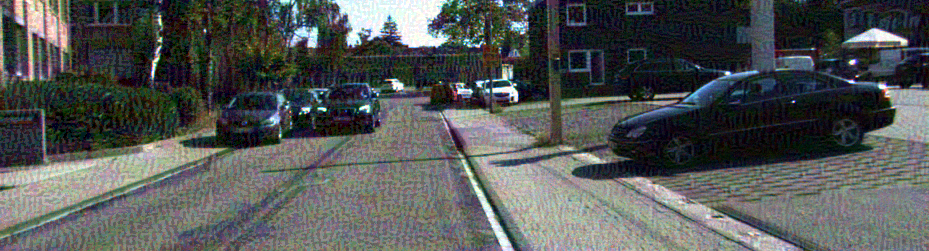}
     \includegraphics[width=0.47\linewidth]{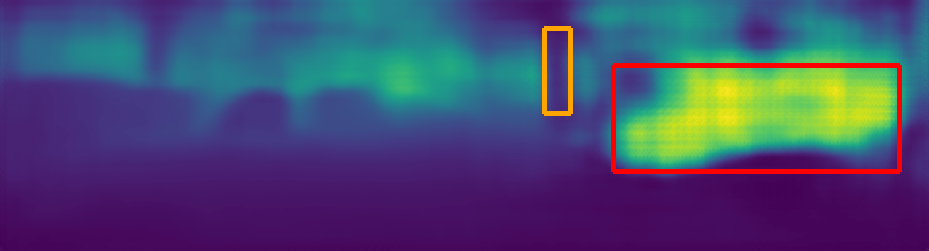}
    }

    \caption{Visualization of targeted attack on 2011\_09\_26/0036/image\_02/0000000050.png of KITTI depth dataset to change the depth result of the target object (the black car in the red boxes) to arbitrary value without changing other part. Besides, the modification to the rgb image is subtle and the depth detail of other parts is preserved well (the pole in orange boxes)}
    \label{fig:exp:specific_depth}
\end{figure}

\textbf{Evaluation metrics.} We adopt root mean squared error (RMSE)~\cite{mal2018sparse} and masked mean depth (MMD) to evaluate the performance on \textbf{non-targeted attack} and \textbf{targeted attack}, respectively.
Specifically, given the ground truth depth $Y^*=\{y*\}$ and predicted depth $Y=\{y\}$, RMSE is computed by $\sqrt{\frac{1}{|Y^*|} \sum_{y^* \in Y^*}\|y^*-y\|^2} $.
The \textbf{larger} RMSE is, the farther predicted depth is from the ground truth depth, which means a better attack performance.
Denoting $M=\{m\}$ as the predefined mask, MMD is computed by $\frac{1}{|M|}\sum_{m \in M}y \cdot m$, which is the mean depth value in the masked area (served as the attack targets in the image).
It represents the average distance of objects to the camera. Note that the predefined $C$ in Eq.~\ref{equ:mask-target} is set to $100$ meters in our experiments, thus the \textbf{larger} MMD is, the better performance the attack method shows (\ie, the results are closer to $C=100$).
The predefined masks is the masks of target objects. How these objects are selected will be explained in the following paragraph.

\textbf{Target Objects.} We select cars, people, and riders with small depth values as the attack target. Firstly, we use Mask-RCNN\cite{he2017mask} to predict the instance segmentation for each image and select the instance mask of cars, people and riders. Then we use the sparse depth ground truth of KITTI dataset to estimate the average depth of each instance. Finally, we select instances whose average depth is smaller than $50$m.

\textbf{Baseline models.} We follow the encoder-decoder framework in~\cite{laina2016deeper} to set up the models in our experiments.
Four different backbones are used as the encoding part, including ResNet-18~\cite{he2016deep}, VGG-16~\cite{simonyan2014very}, ResNet-50 and ResNet-101. For the decoding parts, they are composed of 4 UpConv layers followed by a bilinear upsampling layer in all models \cite{mal2018sparse}.
Note that we refer each model by the backbone architecture. The total training epoch is 20 for all models and we show the baseline performance in Table~\ref{baseline}.

\begin{table}[t]
\small
\begin{center}
\begin{tabular}{c|c|c|c|c}
\hline
Model & ResNet18 & VGG16 & ResNet50 & ResNet101 \\
\hline
RMSE(m) & 4.22 & 4.42 & 4.17 & 4.70 \\
\hline
MMD(m) & 20.76 & 20.57 & 20.82 & 21.34 \\
\hline
\end{tabular}
\end{center}
\caption{Baseline performance of different models.}
\label{baseline}
\end{table}


\textbf{Semantic segmentation model used in multi-task attack.} We utilize the state-of-the-art semantic segmentation model, PSPNet~\cite{zhao2017pyramid}, to provide additional supervision signals.
We use Cityscapes~\cite{cordts2016cityscapes} to train the network.
The total epoch is 100 and random crop (321$\times$929 in our experiment to suit the image size in KITTI) is used during training. When optimizing the universal attack with multi-task strategy, two different networks are used for the depth estimation and semantic segmentation.


\textbf{Universal multi-task attack setting.}
To implement universal multi-task attack, root mean squared error(RMSE) loss is selected as $L^{\text{depth}}$ in Equ.~\ref{equ:multitask_loss} and  $L^{\text{semantic}}$ is the widely used softmax loss~\cite{zhao2017pyramid}.

\textbf{Attack settings.}
We use $L_2$ distance as the loss function $L$.
If not specified intentionally, we set the perturbation constraint  $\epsilon=16$.
For iterative attack methods, we set iteration number to $\min(\epsilon+4,  1.25\epsilon)$ following~\cite{kurakin2016adversarial} and step size $\alpha=1$.
For MI-FGSM, we set $\mu=1$ as~\cite{dong2017boosting} suggests.

\subsection{Single Model Attack}

\label{white_black}

In this experiment, we follow~\cite{dong2017boosting} to report the white-box and black-box attacks in the single-model setting, where all four networks attack each other in both \textbf{non-targeted} and \textbf{targeted} attacks. 
Table~\ref{table:white_black_attack} reports the overall results, where models in each row generate adversarial perturbations that are evaluated on models in each column.

For non-targeted attacks, we are showing the mean error (RMSE) values to indicate the extent of distortion by these attacks.
We can see that for white-box attacks, the estimation errors are up to $10\times$ of the original error in the baseline models (Table~\ref{baseline}).
The black-box attacks are not so effective, but still achieve an average error of up to $4\times$ compared to the baselines.

For targeted attacks, we show the average distance in the estimation results in Table~\ref{table:white_black_attack}. We can see that the estimated distance is up to $4\times$ (from around $20$m to $80$m) farther than the original prediction (Table~\ref{baseline}).
As an illustrative example, Fig.~\ref{fig:intro:traser} presents a case to attack a car nearby.
This adversary phenomenon would be really dangerous if a pedestrian or car is estimated to be $68$ meters away when it is actually only $8$ meters away, in scenarios such as automonous driving.



In addition, Table~\ref{table:white_black_attack} also demonstrates the robustness of different backbones. As shown in the shadow cells, ResNet-50 is the most fragile one in both non-targeted and targeted attacks.
Considering its high performance on clean images, this observation is consistent with the previous claim~\cite{tsipras2018robustness} that models with higher performance are more vulnerable.

%

\subsection{Move Objects to an Arbitrary Distance}

This section shows that the depth value of an object can be manipulated to any number, \ie~move any object to an arbitrary distance.
Fig.~\ref{fig:exp:specific_depth} displays a sample attack result. The example image is 2011\_09\_26/0036/image\_02/0000000050.png of the KITTI depth dataset. The attack target is the black car on the right side of the image (red box in Fig~\ref{fig:exp:specific_depth:raw}) and the original average depth of the target vehicle is $11.4$m. After attack, the average depth of the target object can achieve a minimum of $4.4$m and a maximum of $80.8$m. This means the attacker can not only pull the target closer (Fig.~\ref{fig:exp:specific_depth:4m}, from $11.4$m to $4.4$m), but also push it further to nearly arbitrary distance (Fig.~\ref{fig:exp:specific_depth:30m} to Fig.~\ref{fig:exp:specific_depth:68m}, from $11.4$m to a maximum of $80.8$m).

Note that this attack is concealed very well. On the one hand, the operation on the RGB image is very subtle (the left column of Fig.~\ref{fig:exp:specific_depth}). On the other hand, the depth of other parts of the image is barely changed. In Fig.~\ref{fig:exp:specific_depth}, the depth detail of the pole is preserved well (orange boxes in Fig~\ref{fig:exp:specific_depth}) .

\begin{figure}[h]

	\centering
    \subfigure[{\scriptsize Image}]
    {
    \label{fig:exp:universal:rgb}
    \includegraphics[width=0.46\linewidth]{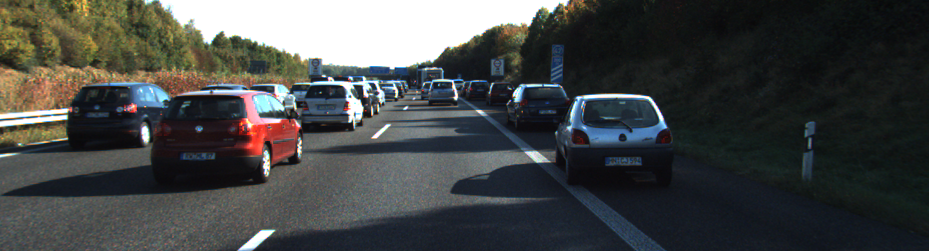}
    \includegraphics[width=0.46\linewidth]{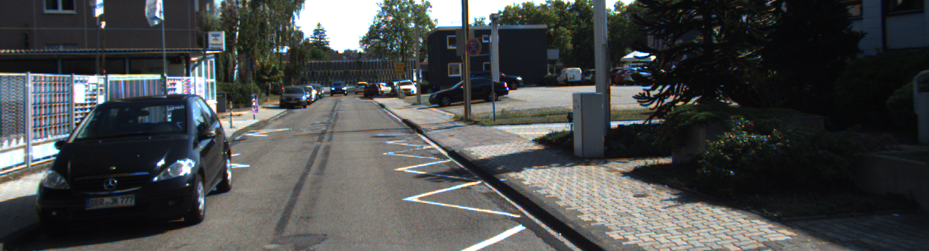}
    }
    
    \subfigure[{\scriptsize Image + UAP}]
    {
    \label{fig:exp:universal:rgb_adv}
    \includegraphics[width=0.46\linewidth]{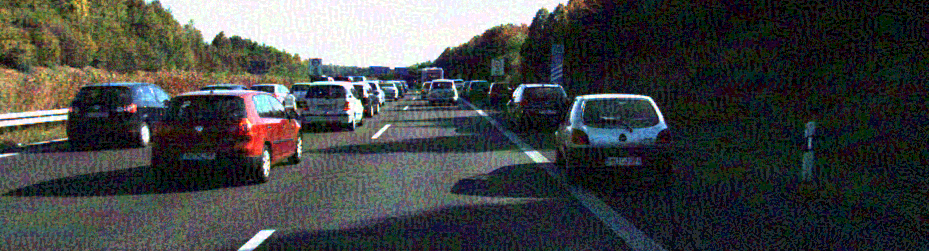}
    \includegraphics[width=0.46\linewidth]{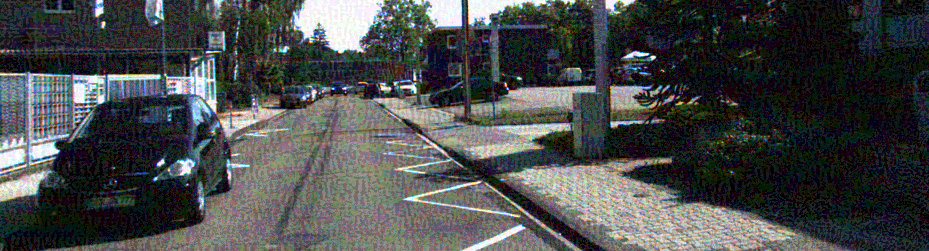}
    }

    \subfigure[{\scriptsize Depth Prediction}]{
    \label{fig:exp:universal:depth}
    \includegraphics[width=0.46\linewidth]{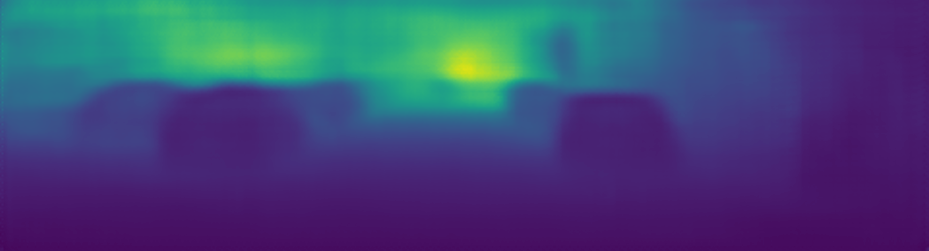}
    \includegraphics[width=0.46\linewidth]{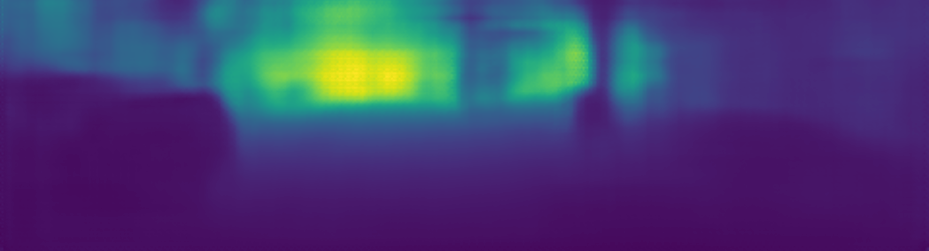}
    }
    
    \subfigure[{\scriptsize Adversarial Depth  Prediction}]{
    \label{fig:exp:universal:depth_adv}
    \includegraphics[width=0.46\linewidth]{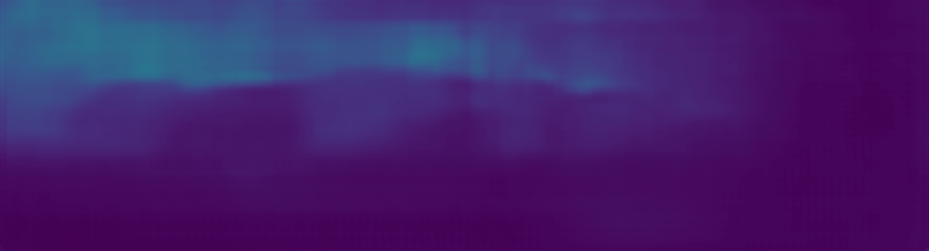}
    \includegraphics[width=0.46\linewidth]{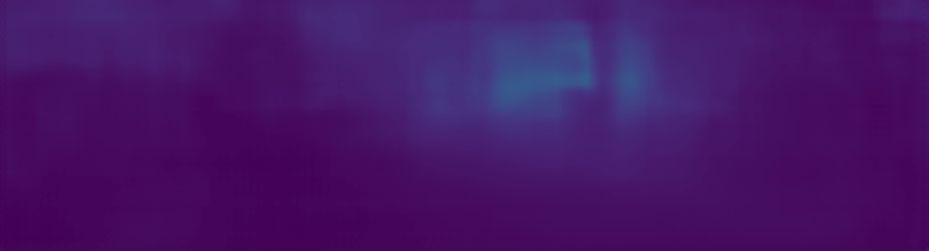}
    }

    \subfigure[{\scriptsize Semantic Segmentation}]{
    \label{fig:exp:universal:semantic}
    \includegraphics[width=0.46\linewidth]{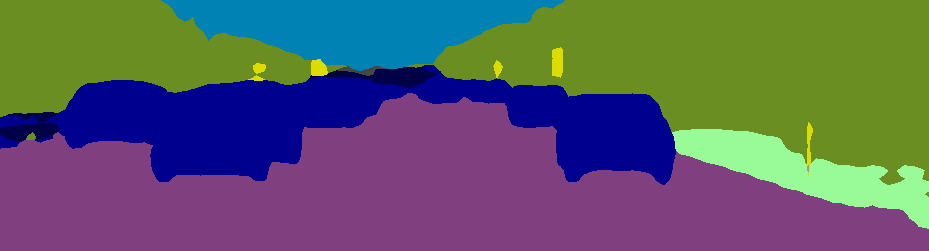}
    \includegraphics[width=0.46\linewidth]{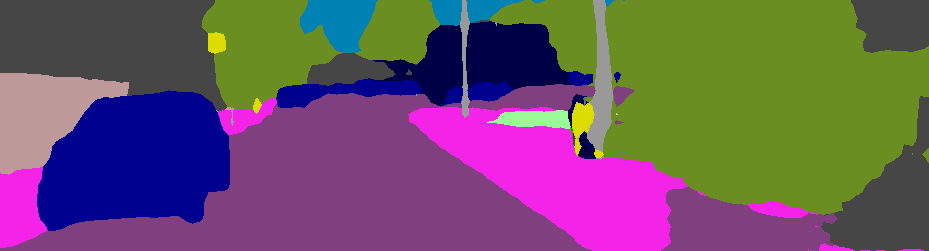}
    }
    
    \subfigure[{\scriptsize Adversarial Semantic Segmentation}]{
    \label{fig:exp:universal:semantic_adv}
    \includegraphics[width=0.46\linewidth]{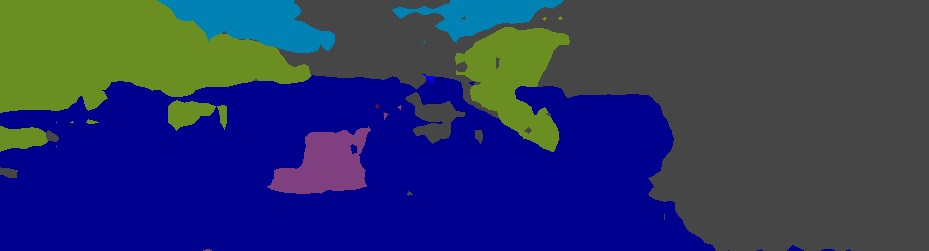}
    \includegraphics[width=0.46\linewidth]{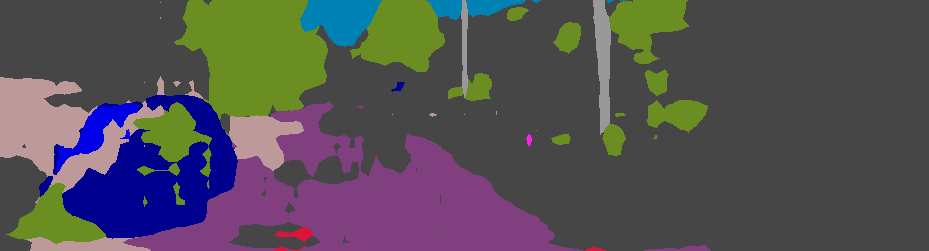}
    }

    \vspace{-2ex}
    \caption{Visualization of the effect of the universal perturbation. 
    The left and right represents two different images.
    }
    \label{fig:exp:universal}

\end{figure}

\begin{table}[h]
\begin{center}
\footnotesize
\setlength{\tabcolsep}{5.0pt}
\begin{tabular}{c|c|c|c|c}
\hline
Model & Method & ResNet-18 & VGG-16 & ResNet-50 \\
\hline
\multirow{4}*{Single-task} & FGSM & 5.967 & 6.784	& 6.18   \\
 & I-FGSM & 6.108 & 6.877 & 6.464  \\
 & MI-FGSM & \textbf{7.173} & \textbf{7.503} & \textbf{7.257}  \\
\hline
\multirow{4}*{Multi-task} & FGSM & 6.693 & 8.098	& 8.197  \\
 & I-FGSM & 5.566 & 6.021 & 5.863  \\
 & MI-FGSM & \textbf{9.378} & \textbf{8.582} & \textbf{9.757}  \\
\hline

\end{tabular}
\end{center}
\caption{ The average depth errors (RMSE) of universal adversarial attacks with respect to different attack methods and settings. Single-task represents utilizing only non-targeted depth information and multi-task means adopting information of both depth and segmentation.}
\label{table:multitask}
\end{table}

\subsection{Universal Attack}

In this experiment, we perform the universal adversarial attack in both \textbf{multi-task} and single-task settings.
We set $w_{\text{depth}}=0.5$, $w_{\text{semantic}}=0.5$ for multi-task setting and $w_{\text{depth}}=1$, $w_{\text{semantic}}=0$ for single-task in Eq.~\ref{equ:multitask}.
Table~\ref{table:multitask} reports the white-box attack result, where
 single-task denotes to only use depth information  and multi-task represents to leverage multiple supervision signals.
As shown in Table~\ref{table:multitask}, with the help of the multi-task strategy, the results of FGSM and MI-FGSM are improved significantly compared to single-task.
This demonstrates that more comprehensive supervision signals do benefit the universal adversarial perturbation generation.
The reason may be that the high-level information of semantic segmentation enriches supervision signal of low-level depth information, thus reinforces attack effect on universal adversarial perturbation.
Fig.~\ref{fig:exp:universal} displays an example of the universal attack, where the results of both depth estimation and semantic segmentation are distorted dramatically.

\section{Conclusion}

In this paper, we present to the best of our knowledge the first systematic investigation of adversarial attacks on monocular depth estimation.
We propose three types of attack, \ie~non-targeted attacks, targeted attacks and universal attacks and adapt three state-of-the-art methods to perform these attacks.
Muti-task setting enriches supervision signals by adopting high-level information for universal attacks.
Extensive experiments demonstrate the vulnerability of depth estimation models and the effectiveness of multi-task strategy compared to different methods.
In particular, we demonstrate the targeted attack on certain objects is able to twist depth estimation up to an average of $4\times$ from the ground truth.
Our work reveals the severe impact of adversarial attacks on depth estimation, as these attacks may result in grave security concerns if applied in cases such as autonomous driving.
We hope this work can provide a baseline and guidance for adversarial attacks research on monocular depth estimation.

\clearpage
{\small
\bibliographystyle{IEEEtran}
\bibliography{IEEEexample}
}

\end{document}